\documentclass[conference]{IEEEtran}

\IEEEoverridecommandlockouts
\usepackage{cite}
\usepackage{amsmath,amssymb,amsfonts}
\usepackage{algorithmic}
\usepackage{graphicx}
\usepackage{textcomp}
\usepackage{xcolor}
\usepackage{tikz}
\usetikzlibrary{graphs}
\usepackage{booktabs}
\usepackage{csquotes}
\usepackage{hyperref}  
\usepackage{array} 
\usepackage{subfig}

\usepackage{balance}

\renewcommand{\O}[1]{\ensuremath{\mathsf{#1}}} 

\usepackage[ruled,vlined]{algorithm2e}
\SetKwInOut{Input}{Input}
\SetKwInOut{Output}{Output}
\SetKwProg{Fn}{Function}{}{}

\def\BibTeX{{\rm B\kern-.05em{\sc i\kern-.025em b}\kern-.08em
    T\kern-.1667em\lower.7ex\hbox{E}\kern-.125emX}}

\makeatletter
\newcommand{\linebreakand}{%
  \end{@IEEEauthorhalign}
  \hfill\mbox{}\par
  \mbox{}\hfill\begin{@IEEEauthorhalign}
}
\makeatother
    
\begin{document}

\newcommand\independent{\protect\mathpalette{\protect\independenT}{\perp}}
\def\independenT#1#2{\mathrel{\rlap{$#1#2$}\mkern2mu{#1#2}}}

\newtheorem{definition}{Definition}[section]


\title{Optimal Transport for Efficient, Unsupervised Anomaly Detection on Industrial Data
\thanks{\copyright\ 2024 IEEE. Personal use of this material is permitted. Permission from IEEE must be obtained for all other uses, in any current or future media, including reprinting/republishing this material for advertising or promotional purposes, creating new collective works, for resale or redistribution to servers or lists, or reuse of any copyrighted component of this work in other works. This is the accepted manuscript of A. Langbridge, F. O'Donncha, J. T. Rayfield and B. Eck, ``Optimal Transport for Efficient, Unsupervised Anomaly Detection on Industrial Data,'' in \textit{2024 IEEE International Conference on Big Data (BigData)}, Washington, DC, USA, 2024, pp. 2142--2151, doi: \url{https://doi.org/10.1109/BigData62323.2024.10825081}.}
}

\author{\IEEEauthorblockN{Abigail Langbridge}
\IEEEauthorblockA{
\textit{Dyson School of Design Engineering} \\
\textit{Imperial College London}\textsuperscript{*}\thanks{\textsuperscript{*}Work completed while at IBM Research Europe}\\
United Kingdom \\
\url{al4518@ic.ac.uk}}
\and
\IEEEauthorblockN{Fearghal O'Donncha}
\IEEEauthorblockA{\textit{AI for Sustainable Industry} \\
\textit{IBM Research Europe}\\
Dublin, Ireland}
\and
\IEEEauthorblockN{James T Rayfield}
\IEEEauthorblockA{
\textit{IBM T.J. Watson Research Center} \\
\textit{IBM Research}\\
NY, USA}
\linebreakand
\IEEEauthorblockN{Bradley Eck}
\IEEEauthorblockA{\textit{AI for Sustainable Industry} \\
\textit{IBM Research Europe}\\
Dublin, Ireland}
}

\maketitle











\begin{abstract}
Effective anomaly detection frameworks are a central pillar of the Industry 4.0 paradigm. In this paper, we introduce an Optimal Transport (OT)-based framework for anomaly detection, designed to detect deviations from normal behaviour in time-series sensor data. The OT-based method requires minimal user input and adapts to real-time data without the need for labelled training data. Our method effectively addresses existing limitations related to data labelling, generalisability, and scalability, demonstrating resilience against short-term fluctuations, noise, and data gaps — common challenges in industrial environments. Additionally, our method provides counterfactual explanations improving the auditability of the approach when deployed in industrial settings.

The proposed method learns the mapping between normal and observed operating conditions through a sliding reference window that adapts to the dynamicity of the data.  We evaluate our approach on three industrial datasets, from shipping, industrial HVAC systems, and publicly available benchmark data.
The method was highly effective in identifying anomalies and reducing false positives, outperforming traditional methods, while maintaining computational efficiency and ease of configuration.

\end{abstract}


\section{Introduction}
Asset performance management (APM) is critical in modern industrial contexts due to increasing automation and reliance on high-cost equipment, necessitating well-planned maintenance to optimize efficiency and reduce costs \cite{zio2013evaluating}. Many firms still rely on calendar-based maintenance due to its simplicity, but this can lead to the waste of significant remaining useful life and increase the risk of unexpected breakdowns \cite{arunraj2007risk}.  Advances in monitoring and analysis have promoted the adoption of condition-based maintenance (CBM), which promises a more effective and targeted scheduling framework. The goal of CBM is to enhance equipment availability and efficiency, control deterioration rates, ensure safe and environmentally sound operations, and minimise total operating costs. A proactive CBM approach can yield substantial cost savings, including up to a 12--20\% reduction in maintenance costs and a 14\% reduction in safety risks\cite{daoudi2023machine}.

\begin{figure}[t]%
    \centering
    \includegraphics[width=\columnwidth]{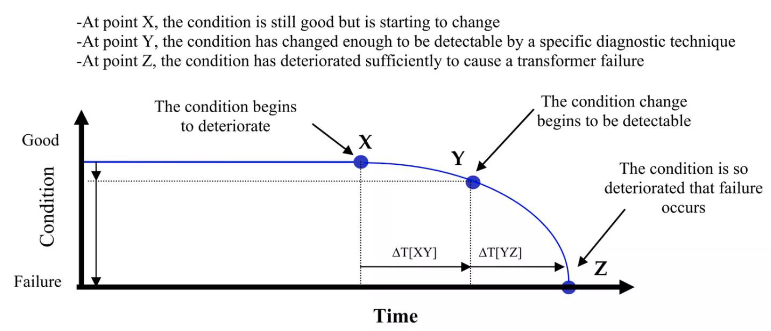}
    \caption{P-F Curve illustrating the progression from potential failure (P) to functional failure (F) in industrial assets. The curve highlights key detection points for condition monitoring and the opportunity window for predictive maintenance interventions to prevent total asset failure}%
    \label{fig:PFcurve}
\end{figure}

The effective implementation of CBM depends on accurately characterising the deterioration process and failure severity, which are complicated by factors such as imperfect condition monitoring and variability in deterioration thresholds. Figure \ref{fig:PFcurve} presents an idealised P-F curve representing the timeline from the point where a potential failure (P) is first detectable in an asset to the point of functional failure (F). The objective of condition monitoring is to identify the onset of failure behaviour prior to the asset suffering deterioration. Anomaly detection plays a crucial role in this framework by identifying early signs of equipment failure or inefficiency, and allowing for timely interventions.

Anomaly detection in industrial asset management presents several significant challenges. These include the high variability in operating conditions, such as fluctuating loads, temperatures, and environmental factors, which complicate the identification of true anomalies. Additionally, a scarcity or complete lack of labelled datasets for failure events, limits the value supervised learning approaches. The complex behaviour of assets, driven by the interactions between various components, further complicates anomaly detection. Noisy, incomplete, or miscalibrated sensor data introduces additional challenges, compounded by the need for real-time processing to avoid operational disruptions. Interpreting detection results accurately and scaling detection models across diverse assets with varying operational contexts also remain critical challenges.

In practical customer settings, the requirements for anomaly detection solutions can be summarised as follows:
\begin{itemize}
    \item \textbf{Automation:} Many industrial operations lack in-house data science expertise, making it essential to have an automated, out-of-the-box anomaly detection solution that can be easily deployed to IoT data streams and configured with minimal effort. Additionally, many organisations do not have access to labelled data for model training, further underscoring the need for solutions that can operate effectively in unsupervised or semi-supervised environments. 
    \item \textbf{Explainability:} Customers need the capability to diagnose the root cause of an anomaly, with clear explanations for the system's decisions or recommendations, detailing the reasoning behind them. Moreover, compliance with safety regulations often mandates the ability to audit past decisions, enabling users or auditors to trace and review the factors and data that influenced a particular decision. 
    \item \textbf{Scalability:} Modern industrial environments typically encompass numerous assets distributed across multiple buildings, sites, or locations, each with distinct environmental characteristics and differing levels of sensor data monitoring. Anomaly detection routines must be scalable, functioning effectively even in conditions where data is sparse or unevenly distributed.
\end{itemize}

In this paper, we present an Optimal Transport (OT)-based framework for unsupervised anomaly detection on industrial data, and demonstrate its effectiveness on three use cases encompassing benchmark and industrial datasets. The proposed method leverages optimal transport to map relationships between expected and anomalous operating conditions from time-series sensor data. This approach offers several key contributions:

\begin{itemize} 
    \item Ease of Configuration: The method requires minimal configuration — users only need to specify the lengths of reference and test datasets for detection. We utilise a sliding window approach that dynamically updates reference datasets with the most recent data, ensuring adaptability to changing conditions. 
    \item Explainability: Leveraging the learned transport mapping, the framework generates counterfactuals that assist users in evaluating deviations from normal behaviour. This helps diagnose root causes of anomalies and quantify the deviation from expected behaviours, making the method highly transparent and interpretable. We assess the characteristics of the generated counterfactuals using an industry dataset. 
    \item Efficiency with Small Datasets: The method is highly efficient, requiring fewer than 150 data points for effective training across diverse applications, as demonstrated in our experiments. This makes it well-suited for industrial applications with limited data, while minimising computational costs. 
\end{itemize}

\section{Background}
\subsection{Time-Series Anomaly Detection}
Anomaly detection is a well-studied problem across many diverse industries and applications. Despite extensive research, anomaly detection in time-series data remains a challenging problem due to the unique issues it presents, such as noise, missing labels, and concept drift \cite{shaukat2021review}. Many prominent time-series anomaly detection methods are context-specific \cite{meng2023explainable, langone2020interpretable, burnaev2016conformalized}, require significant volumes of labelled data to train \cite{meng2023explainable, pang2021deep, shaukat2021review, langone2020interpretable}, or are not readily interpretable \cite{burnaev2016conformalized, rad2021explainable, ahmad2017unsupervised}. These limitations are exacerbated in industrial contexts by the practical requirements for online detectors that can act on massive volumes of data in near real time \cite{rad2021explainable}. Combined with a high degree of seasonality and drift, it's clear that any successful detector for time series data must be adaptive to changing underlying trends, low-latency, and robust to false-positives caused by noise or data quality issues. Emanating from challenges around computational cost, interpretability, and robustness, the majority of operational anomaly detection systems are still rule-based \cite{khan2024knowledge}.

While many approaches leverage deep learning for anomaly detection \cite{pang2021deep, rad2021explainable}, these have limited utility in online settings due to their computational complexity -- both for training and inference -- and require at least some labelled anomalous instances to train. Additionally, these methods are not interpretable and require some post-hoc explanation.

The most fundamental anomaly detection methods are distance- or density-based approaches such as k-nearest neighbours (KNN). However these methods are sensitive to the distance used, the value of hyperparameters and the local structure of data \cite{ramaswamy2000efficient, breunig2000lof}. The anomaly scores output from these approaches are unbounded, which limits their interpretability. \cite{von2018self}, proposed an extension of distance-based approaches to monitor system degradation. However, this was based on a fixed reference period which limits its application to stationary systems or those with well-labelled data.

Various adaptations of conformal prediction methods have been proposed for anomaly detection \cite{balasubramanian2014conformal}, including KNN-CAD which augments the KNN architecture with an inductive measure of non-conformity which can be interpreted as an \textit{anomaly score} \cite{burnaev2016conformalized}. Similarly, \cite{ahmad2017unsupervised} propose the Numenta HTM detector that uses the prediction error from a temporal encoder/decoder architecture as their distance, and that provides an interpretable anomaly score bounded between 0 and 1. The main drawbacks of these approaches are their computational complexity since the reference measures (the non-conformity measure for KNN-CAD and HTM prediction for Numenta HTM) must be updated with each iteration.



\subsection{Explainable Anomaly Detection}
While many works on anomaly detection focus solely on the accuracy of detections, contemporary works increasingly recognise explainability as a crucial aspect of performance \cite{li2023survey}. In this work, we use the terms explainability and interpretability interchangeably to describe the operator's overall understanding of the model's output.

Many simple models, such as rules, decision trees and linear regression models, are inherently interpretable. However, it's important to note that this interpretability is not unbounded: the size and number of concepts both limit the degree to which explanations are human-comprehensible \cite{lage2019evaluation}.

For models which aren't interpretable, a number of post-hoc explanation tools exist which produce local approximations to black-box models \cite{guidotti2018survey, ribeiro2016should, datta2016algorithmic, amarasinghe2018toward}. Many of these methods are tailored to high-dimensional prediction problems, where feature importance-based explanations are valuable. However, for time-series anomaly detection problems additional information on \textit{why} the given feature is anomalous is valuable. The survey \cite{li2023survey} highlight a shortage of sample-based explanation tools for providing this context.



\subsection{Optimal Transport}
Optimal transport is a powerful and flexible tool for comparing distributions. In the below section we introduce the fundamentals of OT. For a more detailed treatment of the topic, readers are directed to \cite{peyr:19}.

Considering the marginal probability measures $\mu$ and $\nu$ (on domains $\mathbb{X}$ and $\mathbb{Y}$ respectively), we define the set of joint distributions $\pi \in \mathbb{M} (\mathbb{X} \times \mathbb{Y})$ where $\mathbb{M}$ is the set of probability measures on the given product domain. The joint distribution $\pi^*$ is the distribution which (i) has $\mu$ and $\nu$ as its marginals, and (ii) minimizes the expected cost of transporting the `mass' from $\mu$ to $\nu$ under some cost function \(\mathsf{C} : \mathbb{X} \times \mathbb{Y} \rightarrow \mathbb{R}\). More formally, the optimal coupling under Kantorovich \cite{kantorovich1942translocation} is defined:
\begin{equation}
    \pi^* = \arg \inf_{\pi \in \Pi(\mu, \nu)} \iint \mathsf{C}(x,y) \pi(x, y) \; \mathrm{d} x \; \mathrm{d} y
    \label{eq:kantorovich}
\end{equation}
where 
\[
\Pi(\mu, \nu) \equiv \left\{ \pi \in \mathbb{M} (\mathbb{X} \times \mathbb{Y}) : \O{T}_{{x}_\sharp} \pi = \mu,\;\;  \O{T}_{{y}_\sharp} \pi = \nu \right\}.
\]
and $\O{T}$ is the push-forward operator such that $\O{T}_{{x}_\sharp} : \mathbb{M} (\mathbb{X}) \mapsto \mathbb{M} (\mathbb{Y})$ and vice versa.

If $\mathsf{C} \equiv \O{L}_p^p$, $p\in\mathbb{N}^+$, the optimal plan $\pi^*$ induces the Wasserstein-$p$ metric in the space $\mathbb{M}$:
\begin{equation}
    \O{W}_p(\mu, \nu)^p = \inf_{\pi \in \Pi(\mu_0, \mu_1)} \iint \O{C}(x,y) \pi(x, y) \; \mathrm{d} x \; \mathrm{d} y
    \label{eq:wasserstein_dist}
\end{equation}

Indeed, the Wasserstein distance has been used for several works in anomaly detection across many domains, including aerospace \cite{seshadri2022spatial}, particle physics \cite{craig2024exploring}, molecular biology \cite{nguyen2024optimal}, and hyperspectral imaging \cite{huyan2022cluster}. However, these methods rely on domain-specific parameterisations. One notable non-parametric approach is presented in \cite{alaoui2020semi}, where the Wasserstein barycenter of a set of reference signals is used to determine whether some test signal is anomalous. Despite its promise, a major drawback of optimal transport (OT) methods for anomaly detection is their computational complexity. Both classic and regularized OT approaches typically require polynomial time, making them computationally demanding. \cite{rubner2000earth, altschuler2022wasserstein}.

The Wasserstein distance is a powerful tool for inherently interpretable decision-making. The authors of \cite{yu2022explainable, sun2023explainable} present an OT-based matching method that not only provides inherent interpretability but also surpasses state-of-the-art matching techniques on complex, heterogeneous legal datasets. A number of works also use the Wasserstein distance to evaluate the interpretability of data and predictive models \cite{miroshnikov2022wasserstein, chaudhury2024explainable}. The work \cite{you2024discount} uses the machinery of optimal transport to produce counterfactual explanations that respect the underlying distributions of factual instances, rather than considering data points independently.


\section{Method}
\subsection{Anomaly Detection using Optimal Transport}
Consider $k$ sensors measuring some instrumented system at regular time intervals. We define the past data $\mathbf{X} \in \mathbb{R}^{(n_\text{ref} \times k)}$ from $t - n_\text{ref}$ to the present time $t$. We then consider the test data $\mathbf{Y} \in \mathbb{R}^{(n_\text{test} \times k)}$, the subsequent $n_\text{test}$ measurements to $t + n_\text{test}$. Since neither $\mathbf{X}$ nor $\mathbf{Y}$ are labelled, we cannot directly use one or the other to evaluate the presence of anomalies. However, since we can assume that anomalies are sparse compared to normal system behaviour, the time-average of $\mathbf{X}$ can be used as a benchmark of normal system behaviour. Rather than storing and computing averages directly, we use a buffer of historic Wasserstein distances to monitor the relative evolution of $\mathbf{X}$ and $\mathbf{Y}$.



\subsubsection{Wasserstein Distances in One Dimension}
Solving for the Wasserstein distance between one-dimensional measures is straightforward and highly efficient. This motivates our univariate approach, where features are treated as independent data streams for faults to be diagnosed from individually.

Considering a single feature $k$, with $\mathbf{x}_k \in \mathbf{X}$ and $\mathbf{y}_k \in \mathbf{Y}$, we can represent the reference and test data as uniform empirical measures $\mu_k = \frac{1}{n_\text{ref}} \sum_i^{n_\text{ref}} \delta_{x_{i,k}}$ and $\nu_k = \frac{1}{n_\text{test}} \sum_j^{n_\text{test}} \delta_{y_{j,k}}$. However, these are susceptible to small-sample errors where outliers distort the empirical distribution. To reduce the impact of these outliers on the resulting Wasserstein distance, we instead consider the arithmetic mean of sorted sequential observations $x_{i,k} \leq x_{i+1, k} \; \forall \; 1 \leq i \leq n_\text{ref}$ as in \cite{langbridge2024overcoming}, such that:
\begin{equation}
    \hat{\mu_k} = \frac{1}{n_\text{ref} - 1} \sum_i^{n_\text{ref} - 1} \delta_{\hat{x}_{i,k}}, \text{where } \hat{x}_{i,k} \equiv \frac{x_{i,k} + x_{i+1,k}}{2}.
    \label{eq:quantised-dist}
\end{equation}
The definition of $\hat{\nu}_k$ over $\hat{\mathbf{y}}_k$ follows.

Given the cumulative distribution function over $\hat{\mu}_k$
\begin{equation*}
    \O{CDF}_{\hat{\mu}_k}(x) \equiv \int_{-\infty}^x \mathrm{d} \hat{\mu}_k \; \forall x \in \mathbb{R},
\end{equation*}
and its pseudoinverse, the generalised quantile function
\begin{equation*}
    \O{CDF}^{-1}_{\hat{\mu}_k}(r) \equiv \min_x \{ x \in \mathbb{R} : \O{CDF}_{\hat{\mu}_k}(x) \geq r\} \forall r \in [0,1],
\end{equation*}
the 2-Wasserstein distance between $\hat{\mu}_k$ and $\hat{\nu}_k$ can be expressed (\cite{peyr:19}, Remark 2.30):
\begin{equation}
    \O{W}_2(\hat{\mu}_k, \hat{\nu}_k)^2 = \int_0^1 \left| \O{CDF}^{-1}_{\hat{\mu}_k}(r) - \O{CDF}^{-1}_{\hat{\nu}_k}(r) \right|^2 \mathrm{d} r.
    \label{eq:2-wass}
\end{equation}
In practice, since the reference data is larger than the test data, the optimal mapping $\pi^*$ associated with this distance is a monotone rearrangement \cite{peyr:19} (Remark 2.28), where the first $\gamma = \frac{n_\text{ref}-1}{n_\text{test}-1}$ elements of $\hat{\mathbf{x}}_{k}$ are mapped to $\hat{y}_{1,k}$, the subsequent $\gamma$ elements to $\hat{y}_{2,k}$, et cetera. Where $\gamma \notin \mathbb{N}$, mass-splitting may occur, such that $\hat{x}_{i,k}$ is mapped in part to both $\hat{y}_{j,k}$ and $\hat{y}_{j+1,k}$.

\subsubsection{Wasserstein Distance as an Anomaly Score}
The Wasserstein distance $\O{W}_2(\hat{\mu}_k, \hat{\nu}_k)^2$ can be interpreted as a measure of the anomalousness of $\mathbf{y}_k$ with respect to $\mathbf{x}_k$. However, this distance is heavily dependent on $\mathbf{X}$ being an accurate representation of good system behaviour, and is unbounded on $\mathbb{R_+}$ which makes it difficult to interpret. To make our approach more robust, we construct a buffer of distances to act as the \textit{memory} of previous behaviour. The buffer $\mathbf{D} \in \mathbb{R}_+^{n_\text{buf} \times k}$ stores the $n_\text{buf}$ most recent Wasserstein distances $\O{W}_2(\hat{\mu}_k, \hat{\nu}_k)^2$, and informs a dynamic threshold for evaluating anomalous behaviour.


In contrast to many other distance-based anomaly detectors, which use empirical quantiles as detection thresholds \cite{alaoui2020semi, seshadri2022spatial}, we adopt a $t$-statistic-based threshold,
\begin{equation}
    z_k = \frac{\O{W}_2(\hat{\mu}_k, \hat{\nu}_k)^2 - E(\mathbf{D}_k)}{\sigma_{\mathbf{D}_k}}
    \label{eq:zscore}
\end{equation}
where $E(\cdot)$ is the expectation, and $\sigma$ the standard deviation, of the distance buffer $\mathbf{D}$. If $z_k$ is greater than some threshold parameter, by default 3.5, then the period $\mathbf{Y}$ is flagged as anomalous for the feature $k$.
To ensure the persistence of anomaly flags throughout anomalous periods, we freeze the reference distribution $\hat{\mu}_k$ when an anomaly is flagged for that feature. Once the anomalous period ends, the distribution is updated based on the most recent reference data $\mathbf{X}$. This sliding window approach, combined with the dynamic threshold, allows the method to adapt to slow-moving system changes, such as seasonality or concept drift, without flagging them as anomalous.








\subsection{Optimal Transport-Based Counterfactual Explanations}
\label{sec:explainable_ot}
To provide explanations for the generated anomaly flags, we leverage the structure of the optimal mapping $\pi^*$. Since $\pi^*$ is a monotone mapping, there will be at least $\lceil \gamma \rceil$ parents for each $\hat{y}_{j,k}$, which form some continuous subset of $\hat{\mathbf{x}}_k$. The counterfactual for any given $\hat{y}_{j,k}$ corresponds to the range of these parents, and can be interpreted as the expected behaviour of the system which the anomaly violates. This formulation also generalises to queries $y \notin \hat{\mathbf{y}}_k$, provided $y$ is drawn from the same underlying distribution as $\hat{\mathbf{y}}_k$. For $j$ such that $\hat{y}_{j,k} \leq y \leq \hat{y}_{j+1,k}$, the counterfactual of $y$ will be the range of the parents of both $\hat{y}_{j,k}$ and $\hat{y}_{j+1,k}$.
This behaviour is critical for monitoring the system after an anomaly has been flagged.

Using OT to provide explanations ensures that explanations are faithful (since they are generated by the same mechanism as the anomalies themselves) and factual (since explanations must be drawn from the reference data). These two properties are designed to improve operator trust in the detections and their explanations, while providing insight into both the current and expected system operation.




\section{Experiments}\label{sec:experiments}

\begin{table*}[t]
\centering
\caption{Summary of the sensor data analysed over the three case studies: a crude oil tanker ship, chiller data from a Poughkeepsie data centre, and publicly available benchmark data extracted from the Numenta Anomaly Benchmark (\url{https://github.com/numenta/NAB}). Note that missing values include downtime for the HVAC system.}
\begin{tabular}{llrrr}
\toprule
\textbf{Case Study} & \textbf{Feature} & \textbf{Sampling Rate} & \textbf{Num. Values} & \textbf{Missing Values (\%)}\\
\midrule
DT4GS Main Engine & Turbocharger inlet temperature (\textdegree C) & 15 min & 32,286 & 0.04\\
& ME scavenging air pressure (Bar) & 15 min & 28,986 & 0.03\\
\midrule
NAB Benchmark & Office temperature (\textdegree C) & 5 min & 22,696 & 0.0 \\
 & Machine temp (\textdegree C) & 60 min & 7,268 & 0.0 \\
 & Cluster CPU use (\%) & 5 min & 18,051 & 0.0  \\
 & Server CPU use (\%)  & 5 min & 4,033 & 0.0 \\
\midrule
HVAC System & Condenser water flow (L/min) & 15 min & 138,078 & 70.73\\
 & Chiller setpoint (\textdegree C) & 15 min & 138,078 & 46.79\\
 & Supply temperature (\textdegree C) & 15 min & 138,078 & 46.82\\
 & Return temperature (\textdegree C) & 15 min & 138,078 & 46.82\\
\bottomrule
\end{tabular}
\label{tab:data_summary}
\end{table*}

\subsection{Datasets}
\subsubsection{Digital Twin for Green Shipping (DT4GS)}
Shipping is vital to global trade, facilitating the transportation of over 90\% of goods. However, it also contributes to 3\% of global greenhouse gas emissions -- emissions that have increased by 20\% over the past decade \cite{unctad2023review}. Improving the efficiency of the shipping industry is crucial for both economic security and achieving environmental targets. Complicating this challenge is an ageing global fleet. As of early 2023, the average ship age was 22.2 years, with over half the fleet now older than 15 years. Many of these ships are 
too old to retrofit while being too young to scrap \cite{unctad2023review}. Therefore, it is imperative to harness digitalisation and technological advancements to enhance efficiencies for the shipping industry.

A key pillar of improving efficiency is the real-time monitoring of various components and systems on a ship, using sensors and data analytics to assess their performance. The primary goal is to detect early signs of wear, degradation, or malfunction, enabling timely maintenance while preventing unexpected failures \cite{mavrakos2024digital}. Anomaly detection is essential in this process, as it facilitates the early identification of potential issues in machinery or systems before they escalate into significant failures. Crucially, any actionable anomaly detection approach must maintain a near-zero false positive rate to ensure reliability and effectiveness \cite{lopez2023fusing}.

In this study, we utilised real-world data from a 9-year-old, 330-meter crude oil tanker vessel. While data were collected from various ship components, this analysis focuses specifically on sensor streams from the main engine, collected every 15 minutes over a continuous 1-year period. Table \ref{tab:data_summary} outlines the data analysed, which focuses on the thermal conditions, rotational speed, and air pressure within the engine's turbocharger system. These data streams offer valuable insights into the turbocharger's thermal and mechanical performance, as well as the air pressure dynamics within the main engine, allowing for an in-depth analysis of engine efficiency, operational stability, and potential performance anomalies.

\subsubsection{Benchmark Data}
The Numenta Anomaly Benchmark \cite{ahmad2017unsupervised} provides 58 labelled time-series datasets designed to test the performance of online anomaly detection approaches in data-streaming settings. The benchmark datasets, many from real-world applications, have varying sample frequency and degrees of seasonality or concept drift. From this corpus, we select the four real-world datasets with known anomaly causes that most closely align with our application. These are detailed in Table \ref{tab:data_summary}.

\subsubsection{Industrial HVAC systems}
Heating, Ventilation, and Air Conditioning (HVAC) systems are essential for maintaining optimal environmental conditions and are widely used across various industries, including manufacturing, data centres, commercial buildings, and healthcare facilities. However, accurately identifying anomalies in HVAC systems is challenging due to their diverse configurations, complex operations, pronounced seasonality, and the variety of building monitoring systems that track performance \cite{granderson2023labeled}. The goal is to leverage operational data collected from building automation systems, sensors, and meters to automatically detect equipment and control issues, as well as degrading performance in HVAC systems, and to diagnose potential root causes \cite{misra2021handbook}.

In this study, we collected data every 15 minutes over a 4-year period from a data centre in Poughkeepsie, NY, focusing on a 15-year-old closed-loop water-cooled chiller equipped with various IoT sensors, as summarized in Table \ref{tab:data_summary}. The data were sourced from the SkySpark\footnote{SkySpark, SkyFoundry, 2019. \url{https://skyfoundry.com/product}} IoT platform, which continuously gathers information via live connections to the site's building management system. The gathered data have significant periods missing due to a combination of sensor faults and scheduled downtime, which are difficult to distinguish from the collected data. We transform the raw sensor data into a set of key performance indicators (KPIs), including Delta Temperature, Delta Setpoint, and Cooling Load. These KPIs offer a more nuanced understanding of the chiller’s operation cognisant of variations in load. It enables more accurate detection of performance issues and enhances decision-making for maintenance and energy management.


\subsection{Baselines and Evaluation Metrics}
To evaluate the performance of our approach, we consider the F1 score, precision and recall of detections against windowed ground-truth labels. Using windowed labels rather than an instantaneous ground truth allows us to reward both early and late detections within some acceptable period of the instantaneous anomaly.
Where ground-truth labels are unavailable, such as in our shipping case study, we consider additional qualitative metrics based on desirable behaviour, such as the density of detections and the robustness to sensor fluctuations.

We explored two OT-based approaches, distinct in their approach to thresholding, with the aim of reducing false positives:
\begin{itemize}
    \item OT-AD: data were flagged as anomalous if the Wasserstein distance  between the observed data and reference distribution was greater than 3.5 times the standard deviation, as in Equation \ref{eq:zscore}.
    \item OT-AD (Q99): data were flagged as anomalous if the Wasserstein distance exceeded the 99th percentile of the distance buffer.
\end{itemize}

We compare our performance to two unsupervised, online detection methods: KNN-CAD \cite{burnaev2016conformalized}, and Numenta HTM \cite{ahmad2017unsupervised}. These methods were some of the top performers on the NAB benchmark \cite{ahmad2017unsupervised} and both are designed to perform on noisy, real-world data.

We also consider the runtimes of detectors as a proxy for computational complexity, a key deployment consideration. Runtimes are reported based on an 36GB M3 Pro Macbook Pro running Sonoma 14.6 and Python 3.12, using the Cython implementation of the Wasserstein solver from \cite{flamary2021pot}. All results reported for Numenta HTM, including runtimes, are taken from \cite{ahmad2017unsupervised} due to its Python 2 dependency. Code to replicate our experiments will be provided upon acceptance.

Since explanations are difficult to evaluate quantitatively, we focus on a qualitative evaluation considering their faithfulness to the underlying data.
We achieve this using a visual diagnostic approach adapted for time-series data from \cite{goode2021visual}.



\section{Results}

\begin{figure*}[t]
    \centering
    \includegraphics[width=0.9\textwidth]{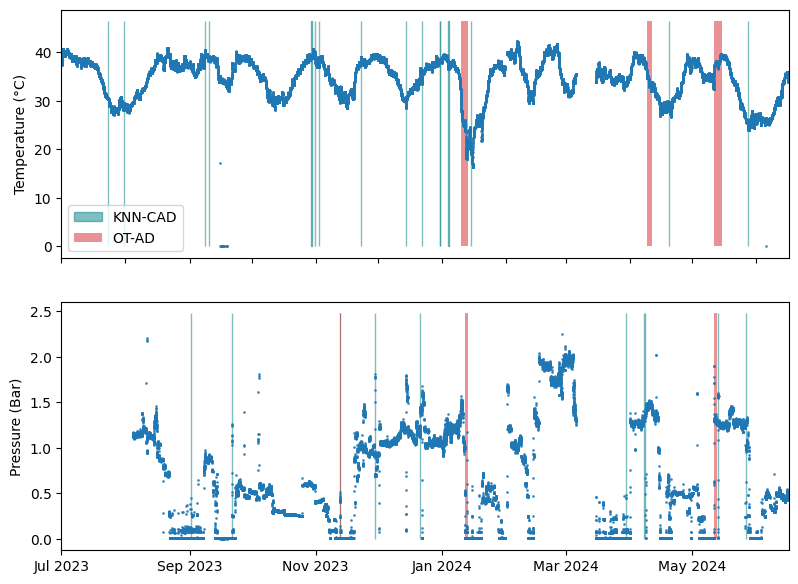}
    \caption{Anomalies detected over one year of engine data from a tanker ship, for the datasets \textit{main engine turbocharger air inlet temperature} (top) and \textit{main engine scavenging air pressure} (bottom). The cyan-highlighted segments represent anomalies identified by the benchmark KNN-CAD method, while the red-highlighted segments indicate anomalies detected by the proposed optimal transport-based approach.}
    \label{fig:dt4gs_results}
\end{figure*}

\subsection{Evaluation on DT4GS Data}
We evaluated our results using one year of engine data. The analysis focused on temperature and pressure data, allowing us to examine anomaly detection across datasets with distinct characteristics:
\begin{itemize}
    \item \textit{Main engine turbocharger air inlet temperature} represents the temperature of the air entering the turbocharger of the main engine. This data can exhibit variability due to changes in engine load, ambient air temperature, and operating conditions.
    \item \textit{Main engine scavenging air pressure} refers to to the pressure of the air supplied to the engine's cylinders during the scavenging process, where exhaust gases are expelled, and fresh air is introduced for the next combustion cycle. Variations in this pressure can indicate changes in engine performance, such as clogged air filters, turbocharger issues, or other mechanical problems affecting air intake.
\end{itemize}

Figure \ref{fig:dt4gs_results} showcases the anomalies detected in both datasets using our OT-based method and the KNN-CAD approach from \cite{burnaev2016conformalized}. For OT-AD, we set the parameters $n_\text{ref}$, $n_\text{test}$, $n_\text{buf}$ all to 48, reflecting the expected anomalous behaviour of persistent faults and resulting in a maximum reporting latency of 12h. For KNN-CAD, we set the detection threshold to 0.99 to reduce the number of detections to only the most extreme anomalies as measured by its conformal model.

For the temperature data, our method identifies three significant anomalous periods in January, April, and May. The first anomaly is triggered by a sharp temperature drop from approximately 34°C to as low as 16°C. Notably, our method detects the anomaly as early as when the temperature falls to 32.3°C, demonstrating its sensitivity and ability to identify anomalies in their early stages.

The comparison in Figure \ref{fig:dt4gs_results} between the OT-AD detections and those from the KNN-CAD implementation highlights key advantages of our approach. 
\begin{itemize} 
    \item The OT method demonstrates greater resilience to short-term fluctuations, which are often the result of sensor errors or data quality issues rather than true anomalies \cite{nunes2023challenges}. In contrast, KNN-CAD mistakenly flags such fluctuations as anomalies, as seen in December 2023 and June 2024. 
    \item The computational efficiency of our OT-AD method is significantly higher. Processing both datasets, each with approximately 30,000 data points, takes only 0.1 seconds for the OT method, compared to 156.1 seconds for KNN-CAD. 
    \item The OT-AD method triggers far fewer anomalous events, identifying just 3 anomalies for both the temperature and pressure datasets, compared to 54 and 45 flagged by KNN-CAD, reflecting a substantial reduction in false positives. 
\end{itemize}


\subsection{Benchmarking on NAB}
In order to achieve the best performance on the benchmark, each method is tuned on each of the four real-world datasets with a grid search conducted to find the parameters leading to the highest F1 score.

The two temperature system datasets are challenging to predict, as reflected by the low F1 scores across benchmark approaches. The optimal thresholds for KNN-CAD are 0.5 and 0.05, and for Numenta HTM are 0.05 for both datasets. The parameters for OT-AD were $n_\text{ref} = 48$, $n_\text{test} = 24$, $n_\text{buf} = 56$ and $n_\text{ref} = 132$, $n_\text{test} = 126$, $n_\text{buf} = 48$ respectively. Notably, the precision of the two OT-based methods are consistently high, representing a significantly lower proportion of false positives compared to the other methods.

For the CPU utilisation data, the optimal parameters for the OT-AD methods are parameters $n_\text{ref} = 128$, $n_\text{test} = 64$, $n_\text{buf} = 64$, and the thresholds for KNN-CAD and Numenta HTM are 0.4 and 0.1 respectively. The change in OT-AD parameters here represents a shift to a different application and different data -- and anomaly -- modalities. Similarly to the temperature data, the OT-AD methods outperform the baselines with respect to precision, and yield similar F1 scores to the best-performing method, Numenta HTM. Notably, the OT-AD approach is more than 10,000 times faster than this method for just a five-percent performance drop. 

Finally, on the request latency data we see a similar trend. The OT-AD methods, with parameters $n_\text{ref} = 18$, $n_\text{test} = 42$, $n_\text{buf} = 42$, achieve perfect precision with an overall F1 score of just under 50\%. The KNN-CAD (threshold 0.5) and Numenta HTM (threshold 0.05) methods suffer from low precision and low recall respectively, which affects their capacity for deployment on this data.


\begin{table}[ht]
    \caption{Performance of our method, KNN-CAD \cite{burnaev2016conformalized}, and Numenta HTM \cite{ahmad2017unsupervised} on four industrial datasets from the Numenta Anomaly Benchmark repository.}
    \centering
    \begin{tabular}{lllll}
    \toprule
    \textbf{Method} & \textbf{Precision} & \textbf{Recall} & \textbf{F1 Score} & \textbf{Runtime (s)}\\
    \midrule
    \multicolumn{5}{l}{\textit{Ambient Temperature System Failure}} \\
    \midrule
    OT-AD & \textbf{0.8} & 0.1322 & 0.227 & \textbf{0.0086} \\
    OT-AD (Q99) & \textbf{0.8} & 0.1322 & 0.227 & 0.0491 \\
    KNN-CAD & 0.1359 & \textbf{0.7617} & 0.2307 & 17.0046 \\
    Numenta HTM & 0.3385 & 0.2107 & \textbf{0.2598} & 82.1171\textsuperscript{*} \\
    \midrule
    \multicolumn{5}{l}{\textit{CPU Utilization ASG Misconfiguration}} \\
    \midrule
    OT-AD & \textbf{0.9003} & 0.3856 & 0.5399 & \textbf{0.0294} \\
    OT-AD (Q99) & 0.8889 & 0.3416 & 0.4935 & 0.077 \\
    KNN-CAD & 0.0939 & \textbf{0.6691} & 0.1646 & 42.9797 \\
    Numenta HTM & 0.7121 & 0.4736 & \textbf{0.5689} & 203.965\textsuperscript{*} \\
    \midrule
    \multicolumn{5}{l}{\textit{EC2 Request Latency System Failure}} \\
    \midrule
    OT-AD & \textbf{1.0} & 0.3121 & \textbf{0.4758} & \textbf{0.0056} \\
    OT-AD (Q99) & \textbf{1.0} & 0.3121 & \textbf{0.4758} & 0.0327 \\
    KNN-CAD & 0.1453 & \textbf{0.7659} & 0.2442 & 8.178 \\
    Numenta HTM & 0.7857 & 0.0954 & 0.1701 & 45.5616\textsuperscript{*} \\
    \midrule
    \multicolumn{5}{l}{\textit{Machine Temperature System Failure}} \\
    \midrule
    OT-AD & \textbf{0.6385} & 0.3902 & \textbf{0.4844} & \textbf{0.0573} \\
    OT-AD (Q99) & 0.2533 & 0.1019 & 0.1453 & 0.1796 \\
    KNN-CAD & 0.1037 & \textbf{0.9383} & 0.1868 & 54.1042 \\
    Numenta HTM & 0.2964 & 0.1204 & 0.1712 & 256.4535\textsuperscript{*} \\
    \bottomrule
    \multicolumn{5}{l}{\textsuperscript{*} Runtimes for Numenta HTM are extrapolated from iteration} \\
    \multicolumn{5}{l}{$\;\;\;$latencies reported in the original paper.}
    \end{tabular}
    \label{tab:NAB_results}
\end{table}

\subsection{Industrial HVAC Data}

\begin{figure*}[t]
    \centering
    \includegraphics[width=0.99\textwidth]{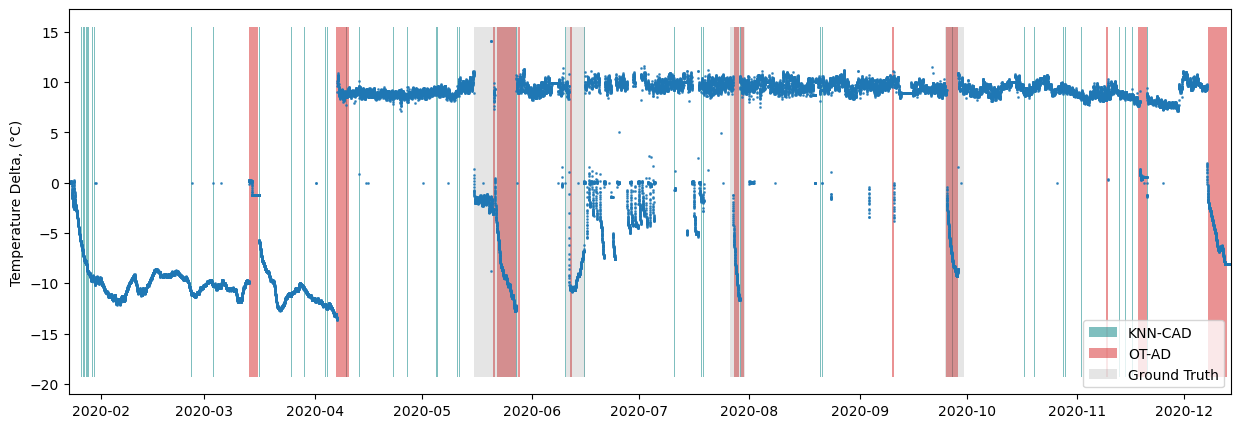}
    \caption{Anomalies detected by KNN-CAD and OT-AD over one year of industrial chiller data. Ground truth anomalies are shown in grey. Note that the KNN-CAD threshold used for plotting is 0.95 (F1 score = 0.03) in order to visually match the proportion of detections to OT-AD.}
    \label{fig:cbm_results}
\end{figure*}

Based on the datastreams in Table \ref{tab:data_summary} we calculate three KPIs to detect anomalies in the chiller system's performance:
\begin{itemize}
    \item \textit{Delta Temperature} is the chiller's supply temperature minus the temperature of water that arrives at the chiller.
    Negative values here correspond to the chiller not doing any cooling work.
    \item \textit{Delta Setpoint} is the system setpoint minus the chiller's supply temperature as a measure of how performant the unit is.
    \item \textit{Cooling Load} is the cooling load on the chiller, estimated as a function of \textit{Delta Temperature} and the flow rate of condenser water through the system.
\end{itemize}


Table \ref{tab:CBM_results} presents the results of OT-AD compared to the baseline KNN-CAD method across these KPIs. The threshold parameter for KNN-CAD is for each feature following a grid search to maximise F1, and the OT-AD parameters are set to $n_\text{ref} = 128$, $n_\text{test} = 24$, $n_\text{buf} = 128$ for all features. The short test window ensures anomalies are detected quickly, while the longer reference window and distance buffer account for the high degree of variation in the system due to noise and operational conditions.

Overall, OT-AD demonstrates a superior balance between precision and recall, leading to a more stable F1 score compared to KNN-CAD. While KNN-CAD excels in recall, capturing more anomalies, it does so at the expense of precision, resulting in a higher rate of false positives. Similarly, the Q99 variation of OT-AD sees slightly lower precision and much lower recall than the original.
The detections of the OT-AD and KNN-CAD (with modified threshold) methods are shown in Figure \ref{fig:cbm_results}. Overall, OT-AD provides the most robust detection, minimising false positives on this complex industrial data compared to KNN-CAD which contains a signficant number of false positives, frequently triggered by noise.




\begin{table}[ht]
    \caption{Performance, stratified by feature, of our method and KNN-CAD \cite{burnaev2016conformalized} on industrial chiller system monitoring data.}
    \centering
    \begin{tabular}{llll}
    \toprule
    \textbf{Method} & \textbf{Precision} & \textbf{Recall} & \textbf{F1 Score} \\ 
    \midrule
    \multicolumn{4}{l}{\textit{Delta Temperature}} \\
    \midrule
    OT-AD & 0.3889 & 0.4574 & \textbf{0.4204} \\
    OT-AD (Q99) & \textbf{0.3931} & 0.2911 & 0.3345 \\
    KNN-CAD & 0.0886 & \textbf{0.7198} & 0.1578 \\
    \midrule
    \multicolumn{4}{l}{\textit{Delta Setpoint}} \\
    \midrule
    OT-AD & \textbf{0.3187} & 0.4581 & \textbf{0.3759} \\
    OT-AD (Q99) & 0.2337 & 0.4581 & 0.3095 \\
    KNN-CAD & 0.1054 & \textbf{0.7713} & 0.1854 \\
    \midrule
    \multicolumn{4}{l}{\textit{Cooling Load}} \\
    \midrule
    OT-AD & 0.4514 & 0.4523 & 0.4519 \\
    OT-AD (Q99) & \textbf{0.6043} & 0.3817 & \textbf{0.4679} \\
    KNN-CAD & 0.0869 & \textbf{0.6615} & 0.1537 \\
    \bottomrule
    \end{tabular}
    \label{tab:CBM_results}
\end{table}

\subsection{Explainable Anomaly Detection}
\begin{figure}[htb]  
    \centering
    \includegraphics[width=\columnwidth]{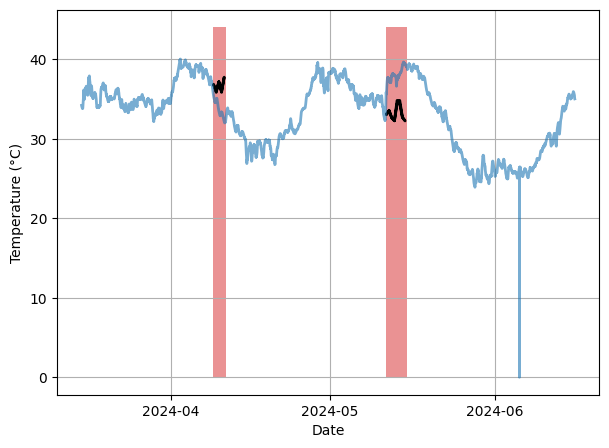}  
    \caption{A section of the DT4GS data in Figure \ref{fig:dt4gs_results}. The highlighted segment in red indicates an anomaly detected by OT-AD, while the black line represents the counterfactual explanation for the last point of each test window.}
    \label{fig:anomaly_counterfactual}
\end{figure}

Figure \ref{fig:anomaly_counterfactual} illustrates the counterfactuals for anomalies detected in the DT4GS case study. For the two flagged events, the counterfactuals reflect how the observations would have behaved according to the reference data.
During the first anomalous period, the temperature drops from approximately 34°C to 32.5°C, while the counterfactuals suggest the expected range should be between 35.9°C and 37.8°C -- in line with the observed behaviour over the previous month. In the second anomalous period, in May 2024, the observed temperature spikes to 39°C, contrasting with the expected range of 32.4°C to 34.9°C which follows the cyclic pattern established after the previous anomalous period. These counterfactuals not only highlight deviations from expected system behaviour but also provide a clear picture of the system’s expected behaviour during the flagged anomalies.

\section{Discussion}\label{sec:discussion}
The results presented on the three industrial datasets highlight the strength of our OT-AD method, particularly on complex systems subject to both operational and environmental fluctuations. The Poughkeepsie case study, in particular, highlights the impact of pronounced seasonal variations on system behaviour. The chiller is switched off during the winter months when ambient temperatures provide sufficient natural cooling, while in the summer, temperatures rise up to 30°C, requiring the chiller to operate at full capacity. This seasonal variation introduces challenges, as any successful anomaly detection framework must distinguish between normal operational patterns and genuine anomalies. Our OT-AD method is able to adapt to shifts within seasons, such as varying daily or weekly cooling demands, as well as long-term wear and degradation, thanks to the adaptive thresholding and sliding reference periods. This approach also proved successful in avoiding the detection of short-term fluctuations in the DT4GS case study, such as those that occurred towards the end of June, which other methods might falsely flag as anomalies. This illustrates the method’s ability to focus on significant deviations while ignoring noise and transient events, providing more actionable insights.

Throughout the studies on labelled data, our OT-AD approach achieves high precision and good F1 scores compared to two benchmark methods. This reduction in false positives is crucial for deployment, where frequent alerts can reduce the percieved severity of anomalies to the detriment of the health of industrial equipment and the safety of operations. The OT-AD method also consistently outperforms the quantile variant OT-AD (Q99), showing that the t-statistic-driven threshold is more performant.

The computational efficiency of the method is another key advantage, highlighting the scalability and real-time applicability of OT-AD for large-scale, continuous monitoring tasks. This adaptability to streaming data makes the method highly relevant for Edge AI applications, where computational resources may be limited and quick decision-making is required. In the context of the shipping industry, where real-time data monitoring is crucial, the evolving trend towards a cloud-Edge continuum enhances the relevance of this approach. These paradigms typically involve a combination of data processing onboard the ship or in the cloud depending on constraints such as operational criticality, security, data volumes, and data transmission costs \cite{odonncha2024towards}. Hence, low cost anomaly detection methods are critical in many settings. 

Another beneficial property of our proposed method is its inherent interpretability, which is critical for industrial applications that prioritise auditability and transparency. Unlike black-box models, which often offer limited interpretability, our method empowers operators to confidently address anomalies by providing counterfactual explanations -- what the system was expected to do based on reference mappings -- rather than relying on abstract metrics such as feature importance. 
The structured nature of optimal transport ensures that these counterfactual explanations are faithful and factual, enabling operators to grasp the underlying causes of the detected anomalies. This transparency is especially vital in industries like manufacturing, shipping, and energy, where safety, efficiency, and regulatory compliance are paramount \cite{ahmed2022artificial}. Similarly, this interpretability extends to the parameter selection step. Since the three key hyperparameters of our OT-AD method have physical meaning in the system to be monitored, operators can utilise their understanding of system operation -- and the anomalies most important to be detected -- to set the parameters appropriately.

\section{Conclusion}\label{sec:conclusion}
In this paper, we present an optimal transport-based method for anomaly detection on unlabelled industrial asset data. By constructing a sliding reference window and monitoring system behaviour with respect to this, our highly-efficient method provides explainable detections that are robust to seasonal trends and concept drift. We evaluate the performance of our approach on a popular anomaly detection benchmark as well as against two industrial datasets from different industries to showcase the consistent performance of the approach across different applications.

Many industrial applications, such as HVAC systems, exhibit pronounced seasonal dynamics. In this study, we used a fixed-length sliding window for anomaly detection throughout the entire period. However, adapting the window size to account for known seasonal variations in data could further enhance the method's robustness and reduce false positives.

Real-time monitoring of false positives and false negatives can help dynamically adjust anomaly score thresholds, ensuring the system remains sensitive to true anomalies without overreacting to noise. This approach can be seamlessly integrated into a rolling window framework, allowing periodic adjustments based on recent performance data.

Agentic large language model (LLM) frameworks could further enhance our anomaly detection method by automating the integration of human feedback. These models can interpret operator input, adjust detection thresholds, and refine parameters in real time, creating a continuous feedback loop that improves accuracy and responsiveness. Additionally, LLMs can help generate auditable records of anomalies and corrective actions, streamlining maintenance processes in industrial applications.

Further work is also planned to investigate the intelligent combination of anomalies detected from different features or KPIs in order to design robust system-level anomaly detection based on known system architectures.

The proposed OT-based anomaly detection framework provides a robust, explainable, and computationally efficient solution for real-time monitoring across diverse industrial applications. Its ability to deliver explainable anomalies and adapt to operational context makes it a valuable tool for modern condition-based monitoring systems, offering asset-specific insights and enhancing decision-making processes for maintenance and operational management. This framework holds significant promise for improving the reliability and efficiency of industrial systems while minimizing false positives and ensuring transparency.


\section*{Acknowledgements}
Research described in this paper has received funding from the European Union’s Horizon Europe Research and Innovation program under grant agreement no. 101056799.



\balance
\bibliographystyle{IEEEtran}
\bibliography{IEEEabrv, bibliography}

\end{document}